\documentclass[preprint,12pt,authoryear]{elsarticle}

\usepackage{amssymb}

\usepackage{lineno}

\usepackage{duckuments}
\usepackage{subcaption}
\usepackage{booktabs}

\journal{Aquacultural Engineering}

\begin{document}

\begin{frontmatter}



\title{Smart Headset, Computer Vision and Machine Learning for Efficient Prawn Farm Management}

\author[act]{Mingze Xi}
\author[sandybay]{Ashfaqur Rahman}
\author[act]{Chuong Nguyen}
\author[bribie]{Stuart Arnold}
\author[sandybay]{John McCulloch}

\affiliation[act]{organization={Data61 CSIRO},
            addressline={Clunies Ross Rd}, 
            city={Acton},
            postcode={2601}, 
            state={ACT},
            country={Australia}}
            
\affiliation[sandybay]{organization={Data61 CSIRO},
            addressline={College Rd, University of Tasmania}, 
            city={Sandy Bay},
            postcode={7005}, 
            state={TAS},
            country={Australia}}

\affiliation[bribie]{organization={Agriculture and Food},
            addressline={North Street}, 
            city={Bribie Island},
            postcode={4507}, 
            state={QLD},
            country={Australia}}

\begin{abstract}
Understanding the growth and distribution of the prawns is critical for optimising the feed and harvest strategies. An inadequate understanding of prawn growth can lead to reduced financial gain, for example, crops are harvested too early. 
The key to maintaining a good understanding of prawn growth is frequent sampling.
However, the most commonly adopted sampling practice, the cast net approach, is unable to sample the prawns at a high frequency as it is expensive and laborious. An alternative approach is to sample prawns from feed trays that farm workers inspect each day. This will allow growth data collection at a high frequency (each day). But measuring prawns manually each day is a laborious task. 
In this article, we propose a new approach that utilises smart glasses, depth camera, computer vision and machine learning to detect prawn distribution and growth from feed trays.
A smart headset was built to allow farmers to collect prawn data while performing daily feed tray checks. A computer vision + machine learning pipeline was developed and demonstrated to detect the growth trends of prawns in 4 prawn ponds over a growing season.

\end{abstract}




\end{frontmatter}


\section{Introduction}
\label{sec:intro}


In prawn farming, continual monitoring of the average weight and size distribution of prawns in a pond is essential to optimise husbandry and harvest strategies. Current best practice, both domestic and international, involves casting a net that catches a sample of up to 100 prawns from the pond. Captured animals are bulk weighed and individually counted to estimate the average weight. This extremely labour-intensive task means that the prawn samples are often only collected from one specific location in the pond and at a low frequency (e.g. once per week). The estimation of the average weight is potentially biased due to the low sampling (net-casting) rate and inconsistency of the number of animals weighed. Erroneous weight estimates can mask sub-optimal growth and underlying pond issues leading to long delays (weeks) before issues are noticed through subsequent measurement. These delays can have a significant economic impact. Prawns are not often weighed individually due to the added time involved, and therefore, size distribution data is not collected.

Frequent data collection on the size of individual animals can provide important information for evaluating growth rates and size distributions, which provide insights into productivity, conditions of the pond and potential yield. This information can help the farm manager predict and avoid unwanted situations. Prawn farm technicians pull up feed trays as part of their daily workflow to understand feed consumption and adjust feed rates. The tray typically captures a good number of prawns because feed is added to the tray to attract the prawns. We aim to take advantage of this practice as this process is more frequent (once/twice daily) than the casting of a net (once every week or fortnight). This is where Smart Headset, Computer Vision (CV) and Machine Learning (ML) can contribute. A smart Headset can be equipped with cameras and farmers can be equipped with these headsets to automatically capture the feed tray images (RGB and depth) hands-free without disturbing their daily operation/workflow. A smart glass (e.g. google glass) can be used by farmers to interact with the headset camera and also display results. Images of the prawns from feed trays can reveal features that can be captured by computer vision methods and converted to size estimates by machine learning algorithms. Automatic and frequent measurements of prawns can provide farmers with valuable insight (not revealed otherwise).

With an aim to provide frequent insight into how the prawns are growing over time, the project aims to develop a pondside smart headset and computer vision-based system to automatically measure prawn size from images acquired using the existing feed tray processes. More precisely, we aim to develop (i) a hands-free smart glass-based field data (RGB and depth image) collection kit to acquire images of feed trays as they are raised, (ii) develop a set of computer vision and machine learning-based methods to estimate prawn size based on those field quality images, and (iii) conduct an analysis of how accurately the measured prawn sizes reveal pond status (e.g. size variation) based on field quality data. 

\section{Materials and methods}
\label{sec:methods}
\begin{figure}
\centering
    \includegraphics[width=0.9\textwidth]{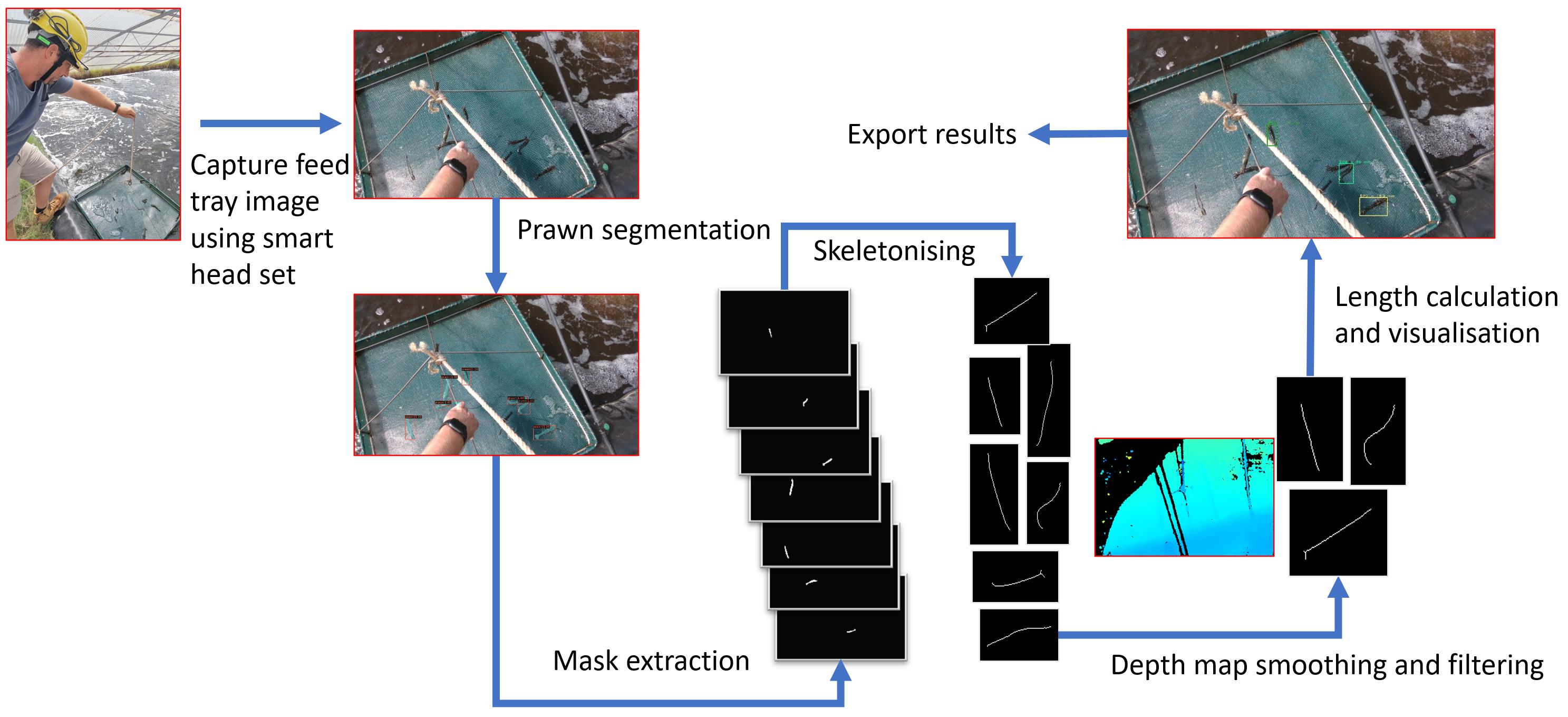}
    \caption{Image Capture, Computer Vision, and Machine Learning pipeline for measuring prawn size.}
    \label{fig:CV-ML-Flow-Chart}
\end{figure}

The overall pipeline of the automated prawn size estimation process is presented in Figure~\ref{fig:CV-ML-Flow-Chart}. Farmers/technicians wear the smart headset and switch the camera on (using a smart glass interface) before pulling out the feed tray from the prawn pond. The images are stored and processed on a unit placed on the back of the headset. Prawns are detected on the tray based on a deep learning (CV+ML) method. A set of image processing methods are then applied to each prawn segment to obtain the centreline of each segment. The centreline in the corresponding depth image (both RGB and depth camera are aligned and synced) is extracted, smoothed (due to noise), and used for computing the prawn size. The prawn size and growth statistics are overlayed on the prawn segments by the smart glass for efficient decision-making. We also experimented with a prawn tracking method to refine the size estimates across frames of a video. Each of these steps is detailed in the following sections.

\subsection{Hands-free prawn data acquisition system}
\label{subsec:headset}
Prawn farm operations are typically performed under great time pressure. For example, on a large commercial farm (e.g., 100+ Hectares), farmers would have to finish examining the feed tray of a pond in only one minute. A hands-free data collection system is essential as farmers usually have both hands occupied, for example, pulling a tray from the pond or holding a water quality sensing device. 

\begin{figure}
    \centering
    \includegraphics[width=.6\textwidth]{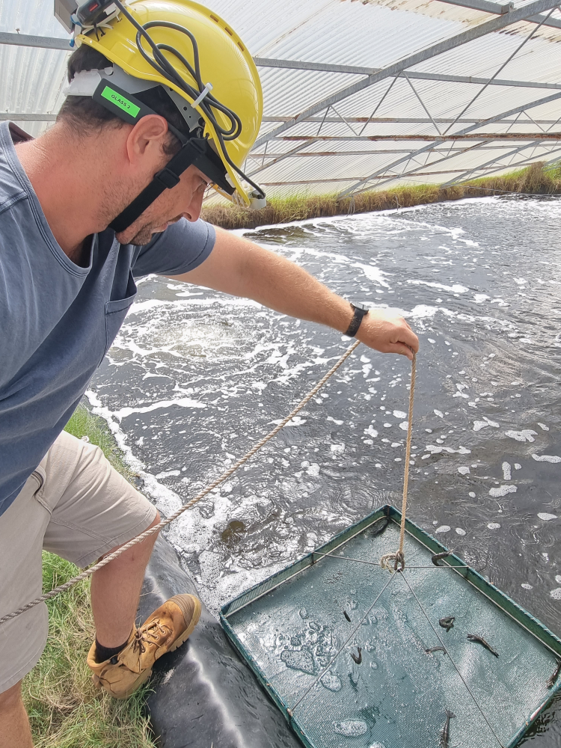}
    \caption{A technician is wearing smart glasses and is using the smart headset to collect prawn data.}
    \label{fig:headset}
    \end{figure}

Unlike many previous works that rely on regular colour cameras~\citep{Thai2021,zhang2022,nguyen2020two} and require tightly controlled lighting conditions~\citep{kesvarakul2017babyshrimp,MOHEBBI2009128}, we use a depth camera as it can provide the real-world coordinates of the prawns required to measure the length accurately. We investigated several alternatives to develop a system that consists of a Google Glass as the client-side viewfinder, a Raspberry Pi 4B as a streaming/recording server, an Intel RealSense D435i depth camera and a power bank for power supply. All hardware components are housed on a hard hat. This smart headset unit (Figure~\ref{fig:headset}) was used successfully in a field environment (see Section~\ref{subsec:prawndata}) to collect video recordings (both RGB and depth images, RGB-D) for processing, training, and testing with computer vision and machine learning methods.

\begin{figure}
\centering
    \includegraphics[width=0.9\textwidth]{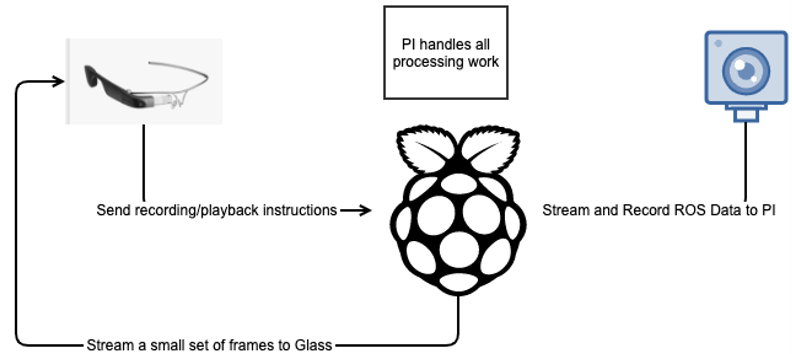}
    \caption{The three main components of the data collection system.}
    \label{fig:software-structure}
\end{figure}

The software implementation contains two main components, i.e., a server app for the Raspberry Pi 4 and a client app for the Google Glass. The components are illustrated in Figure~\ref{fig:software-structure}.

The server app is responsible for recording the video stream and saving the data to a local directory. The client app is responsible for displaying the video stream on Google Glass. The server app is also responsible for sending the data to the server app for processing and training. The server app is also responsible for saving the data to a local directory. The client app is also responsible for displaying the data on Google Glass.

\subsubsection{Software for Raspberry Pi (Server)}
\label{subsubsec:rs4pi}
To make the system field-ready, we have to ensure that the system can set itself up without any human intervention. The common practice requires the user to use a separate computer to remotely control the Raspberry Pi, referred to as the headless mode. However, this is not a good practice for the field environment as farmers are not computer engineers and do not have the required equipment in the field. Our solution is configuring the Raspberry Pi as a server using Nginx (webserver), Flask (python-based web framework) and Gunicorn (web server gateway interface, WSGI), which starts the server-side camera controller app automatically whenever the Pi is turned on. 

The server-side app, called RS4Pi (\textbf{R}eal\textbf{S}ense for Raspberry \textbf{Pi}), uses Flask to handle Glass requests and manage camera setup/stream/record activities, which are implemented using Python, OpenCV and pyrealsense2 library. It also offers basic Pi storage management, such as checking storage usage and removing old recordings. In order to receive commands from Google Glass, we modified the Raspberry Pi network service that turns the Raspberry Pi into a hotspot automatically when the system starts. We then configured Glass to connect to this Wi-Fi network automatically. This way, Google Glass can reach the RS4Pi app and control the camera. 


\subsubsection{Depth Camera Control from Google Glass (Client)}
\label{subsubsec:client}
\begin{figure}[h]
    \centering
    \includegraphics[width=.95\textwidth]{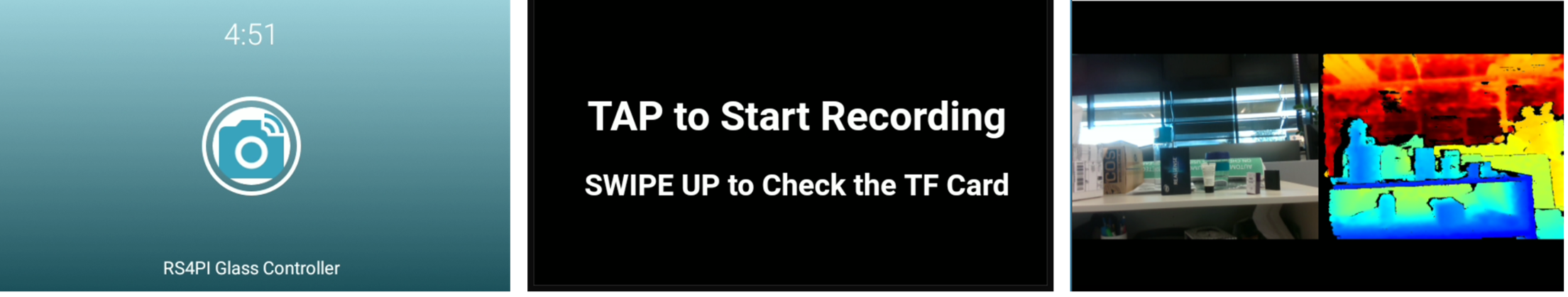}
    \caption{The screenshots of the Android app for controlling the RS camera.}
    \label{fig:glass-view}
\end{figure}

An Android app was developed to allow the user to access the live stream of the RS camera, start/stop recording and check the storage status of the Raspberry Pi (see Figure~\ref{fig:glass-view}). 

\subsection{Headset Assembly (Hardware)}
\label{subsec:headset-assembly}
The hardware, including a power bank, was mounted on a safety hat. With this system, a farmer only needs to put on the Google Glass and the hardhat before leaving the office to do tray checks. 

\begin{figure}[h]
    \centering
    \includegraphics[width=.95\textwidth,trim={0 2cm 8.5cm 0},clip]{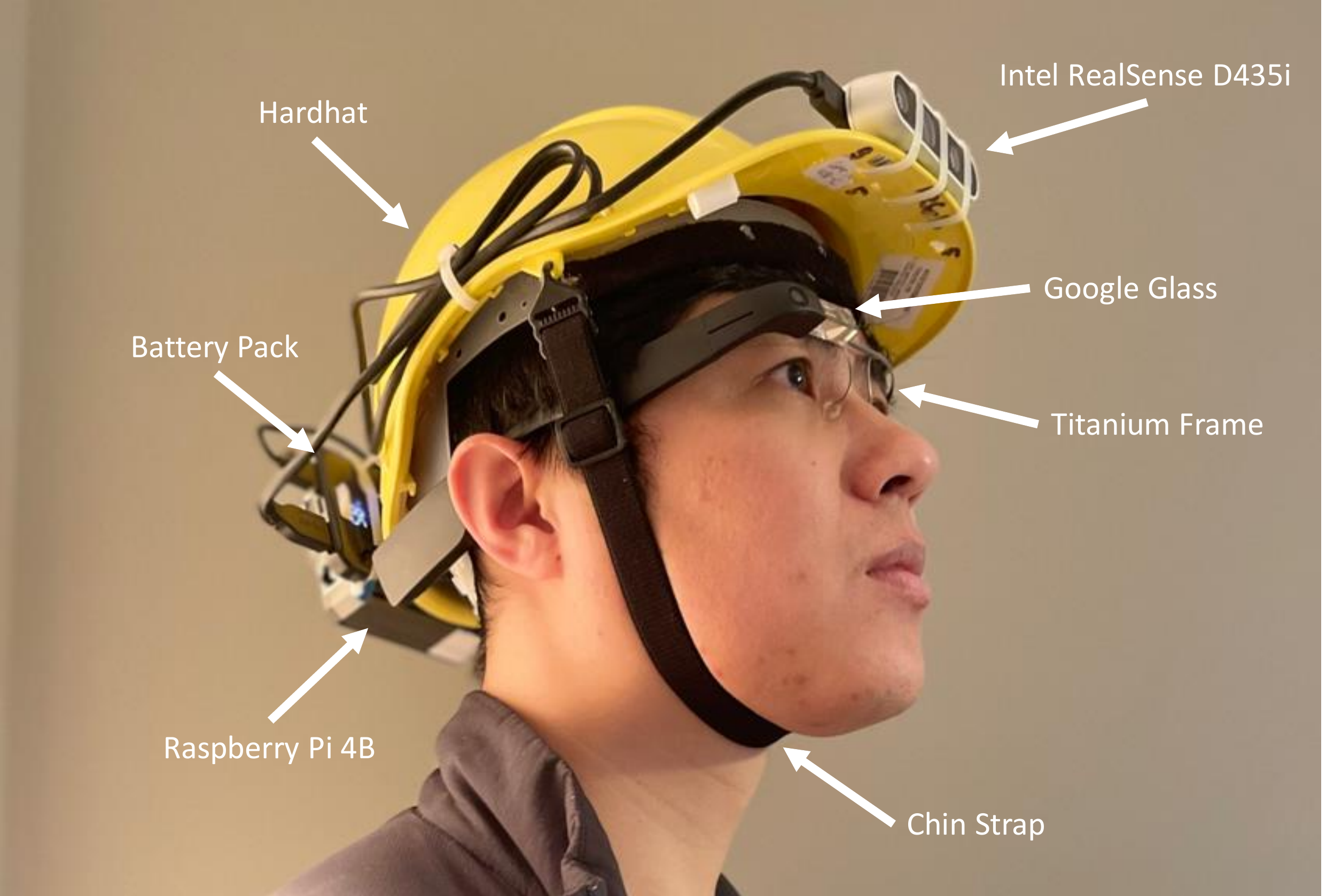}
    \caption{A close look at the helmet. This figure should label different components that have been installed onto the helmet}
    \label{fig:helmet-components}
\end{figure}

\subsection{Prawn dataset}
\label{subsec:prawndata}
\begin{figure}
\centering
    \includegraphics[width=1.0\textwidth]{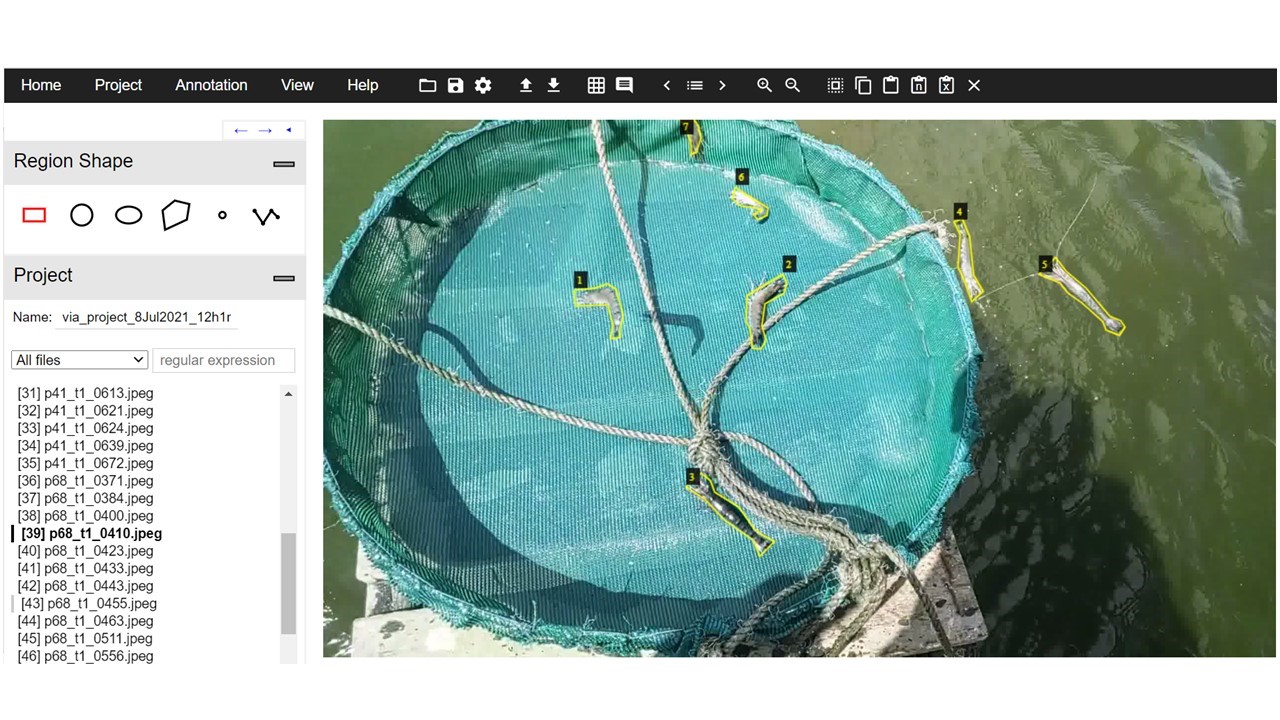}
    \caption{Image annotation done using VGG Image Annotator}
    \label{fig:image-annotation-via}
\end{figure}

One of the key outputs of the project was an annotated data set of prawn images and measurements. The headset was deployed at CSIRO’s Bribie Island aquaculture field station from mid-November to the end of December 2021. Field technicians wore the smart headset whilst conducting feed tray checks at the ponds. We collected field data from four ponds for a period of seven weeks, approximately twice per week. In total, we collected a total of 91 recordings that are stored in Robot Operating System (ROS) Bag format. Each recorded ROS bag includes a colour stream (BGR8, 1280x720, 15FPS), a depth stream (Z16, 1280×720, 15FPS) and two motion streams, including an accelerometer (60FPS) and gyroscope (200FPS). We also hand-measured the lengths of five to six randomly sampled prawns from each tray in the first four weeks of data collection. A total of 4454 prawns from 735 randomly selected RGB images were manually annotated with polygons (Figure~\ref{fig:image-annotation-via}) using VGG Image Annotator~\citep{dutta2019vgg,dutta2016via}. This dataset was later used to train, validate and test the method of using CV+ML to compute prawn size. This dataset is the first of its kind and can be used for future research in both the aquaculture and computer vision/machine learning domains. It is in the process of being made publicly available via the CSIRO Data Access Portal\footnote{CSIRO Data Access Portal - https://data.csiro.au/ [Last Accessed: 27/09/2022]}.

\subsection{Computer vision-based automatic prawn size estimation}
\label{subsec:cv}
This section describes the computer vision pipeline and how it was used to estimate the prawn size.

\subsubsection{Prawn detection}
\label{subsubsec:prawndetection}
Once videos of the feed trays (with prawns on them) are collected, we need a model that can detect the prawns shown on the tray. We are interested in the length and shape of the prawn; hence, it is necessary to get the segmented prawn regions for further image processing. More precisely, we are interested in detecting all the prawns separately (for size measurement), and hence we need instance segmentation~\citep{Hafiz2020} rather than semantic segmentation~\citep{Guo2018}. There are a number of commonly used instance segmentation models, such as Mask R-CNN~\citep{MaskRCNN_2017}, Cascade Mask R-CNN~\citep{cascade_rcnn}, and HTC (Hybrid Task Cascade)~\citep{HTC_2019} with various backbones, including ResNet~\citep{ResNet}, ResNeXt~\citep{ResNext}, SwinT~\citep{SwinT} and DetectoRS~\citep{DetectoRS_2021}.

We trained the models using MMDetection~\citep{mmdetection}, which is a PyTorch-based toolbox that offers a faster training speed. The backbones used in the models were pre-trained on the ImageNet 1K dataset~\citep{ImageNet_2009,ImageNet_2015}.
All models were then trained on the Bribie2021 training dataset with a 2x learning schedule (24 epochs). The models were then benchmarked against the Bribie2021 validation dataset. The best-performing model will be used to generate prawn segmentations, which will be used in subsequent steps to generate centrelines and calculate the physical length of the prawns.

\subsubsection{Prawn skeletonisation}
\label{subsubsec:prawnskeleton}

The next step is to figure out where the “spine” or centreline of each prawn is, i.e. a vector in 2 dimensions describing the prawn’s curving long axis.
The main technique used to produce these centrelines is a type of image processing operation called “skeletonising” or “image thinning”. The Python scikit-image library~\citep{scikit_image} offers three main skeletonising approaches: “Zhang84”~\citep{skel_zhang}, “Lee94”~\citep{skel_lee} and “thin/topological skeleton”. We used the Zhang84 method ($skimage.morphology.skeletonize$) as it is the preferred method for skeletons with few branches (as expected for prawns). As the scikit-image skeletonising algorithms only work with black and white (BW) images, we first converted each RGB image to a BW image using OpenCV. If an image contains multiple detected prawns, we also produce multiple BW images where each image only has one prawn mask. This eliminates the situation where multiple prawns overlap, which will be incorrectly treated as a single animal (skeleton). This also allows us to easily link the computed centreline to a bounding box and a tracker ID produced by a tracking algorithm. The output of the skeletonising algorithm (the centreline) is represented as a sequence of adjacent pixels on the image.

\begin{figure}[h]
    \centering
    \begin{subfigure}{0.45\textwidth}
        \includegraphics[width=\textwidth]{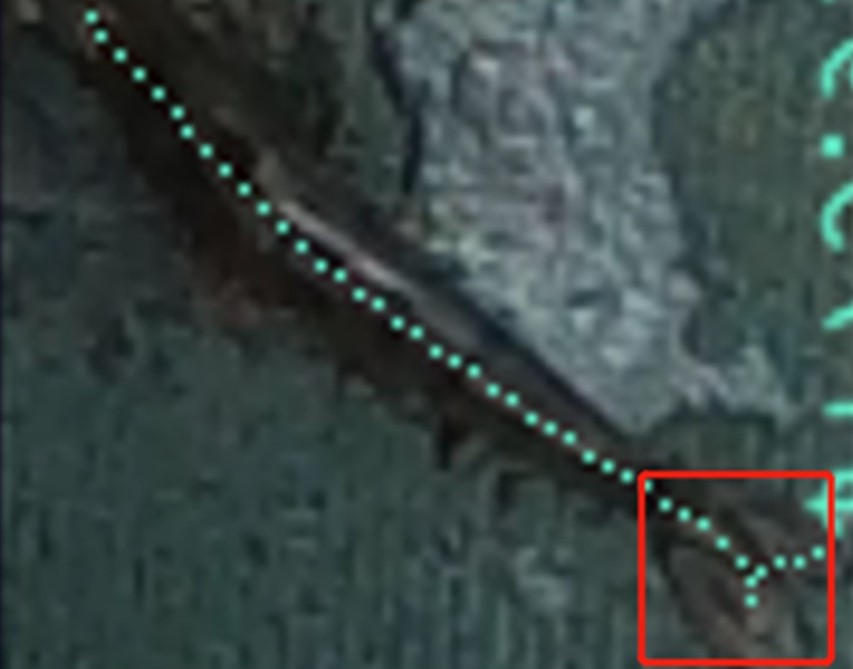}
        \caption{An example centreline that splits around the tail.}
        \label{fig:centreline_p1}
    \end{subfigure}
    \hfill
    \begin{subfigure}{0.45\textwidth}
        \includegraphics[width=.9\textwidth]{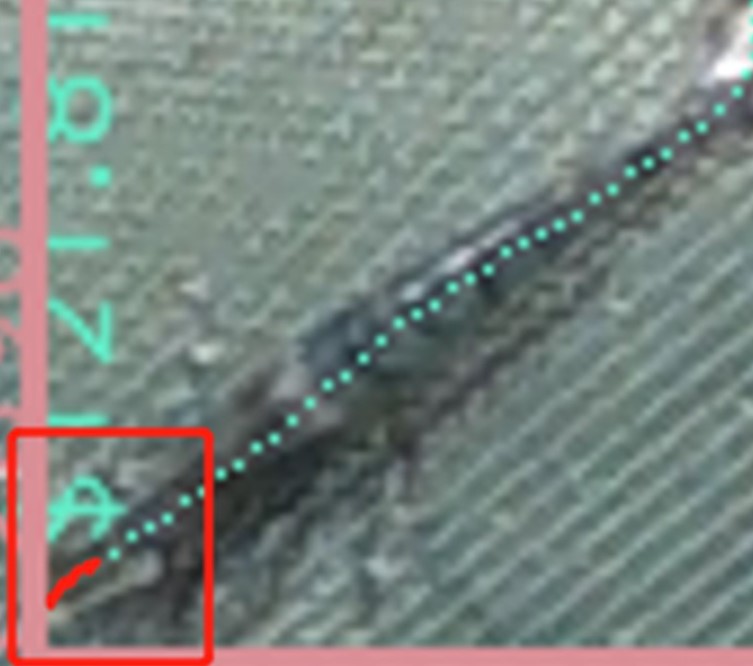}
        \caption{An example of underestimation bias.}
        \label{fig:centreline_p2}
    \end{subfigure}
    \caption{Examples of inaccurate centreline calculations.}
\end{figure}

While the centreline calculation worked well with most detected prawns, we discovered two main issues.
\begin{itemize}
    \item One issue is that the centreline sometimes splits into two branches around the tail (Figure~\ref{fig:centreline_p1}), which leads to a slight overestimation of the length of the prawn. This is inevitable as prawns’ tails naturally split at the end. Future work can investigate using image processing or machine learning algorithms to rectify the tail splits to produce a smoother line without branches.
    \item A second issue is an under-estimation bias introduced by the skeletonising algorithm: the centreline does not always have one end at the tip of the head and the other end at the tip of the tail (Figure~\ref{fig:centreline_p2}). Based on manual examination of a small set of samples, we observed that the bias is minimal, perhaps resulting in under-estimation of the prawn length by $<5\%$.
\end{itemize}

\subsubsection{Prawn length estimation from depth camera}
\label{subsubsec:mlforlength}

An important part of the project is calculating the length of the prawn by utilising depth information. The Intel RealSense D435i depth camera uses two infrared cameras to generate the depth map. By aligning the depth image to the colour image, we could de-project the 2D pixels that form the centreline into 3-dimensional space. This allowed us to calculate the physical distance between any pair of pixels directly. Applying the calculation on the prawn centreline gives us the actual length of the prawn.

The biggest source of error in these calculations is the quality of the depth map. All data collected in this project are from an uncontrolled outdoor environment. Unlike an indoor environment where lighting can be easily controlled, the field environment, including weather conditions and human operations, is entirely unpredictable. This means some depth maps can be extremely noisy. For example, water left on the tray can cause strong reflections on a sunny day, which tends to result in poor depth maps.

The first step we took to mitigate this issue was applying multiple filters before calculating the prawn length. We first down-sampled the pixels that a centreline contains, then excluded invalid pixels (those with no depth value). We then applied another filter to remove pixels on a distorted depth map, for example, where a pixel coordinate was inconsistent compared to the rest of the pixels. After applying all the filters, we calculate the proportion of the total downsampled pixels that are valid and reject a centreline as a measurement of a prawn if this proportion falls below $95\%$. The $95\%$ threshold is an extremely strict rule, which could be tuned down with further fine-tuned depth-map post-processing algorithms, such as temporal filtering, edge preserving and spatial hole-filling.

With all the filters in place, we were able to calculate the lengths of the prawns along a less bumpy reconstructed 3D centreline. However, noise still remained. To further improved the accuracy, we applied a set of smoothing techniques. The main focus here was smoothing in the z-dimension (depth), which directly affects the length estimation. There are two sources of noise in the z-dimension: outliers and missing values (i.e. pixels with no depth information). We first detected extreme outliers in the z-dimension and replaced them with missing values. We then obtained the number of missing segments (a missing segment represents a continuous sequence of missing values) along the centreline. Each of these missing segments was interpolated based on depth values before and after the segment. We also noticed that the computed centrelines could exhibit small zigzags, which are caused by the segmentation and skeletonising algorithm. These lead to a minor overestimation of the prawn’s length. To deal with this situation, we applied 2nd order polynomial fitting to both the x and y coordinates of the centreline pixels. After this process, we obtained a smooth centreline in the three-dimensional space. We computed the length based on the summation of Euclidian distance between successive points along the centreline based on the smoothed 3-dimensional coordinates.

\begin{figure}
\centering
    \includegraphics[width=.95\textwidth]{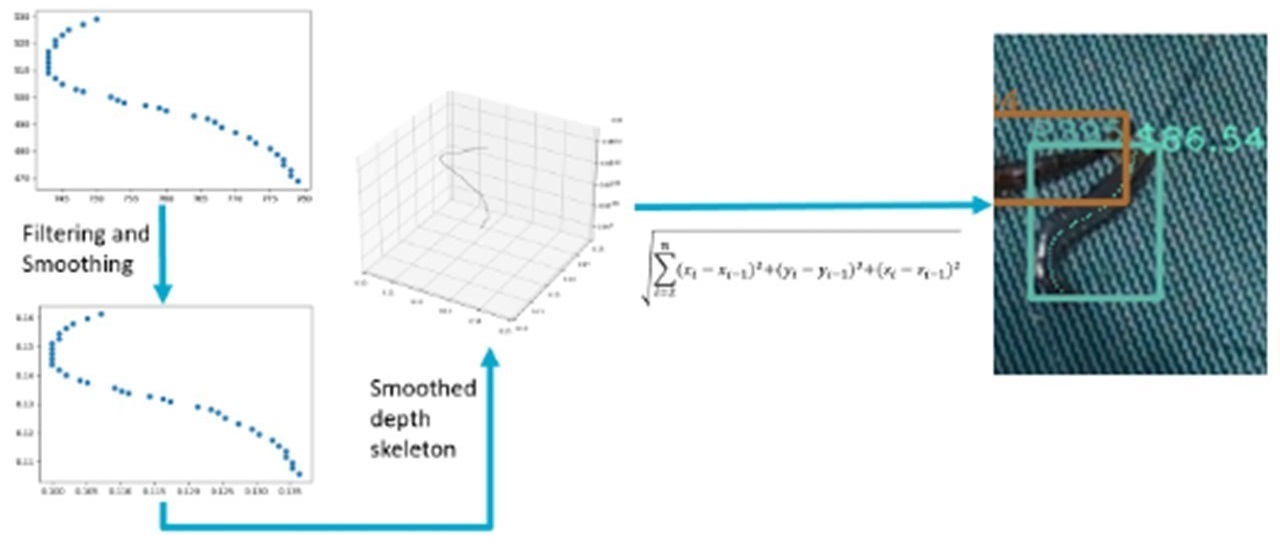}
    \caption{Steps to deal with noise in field quality images and compute length.}
    \label{fig:centrelineCalc}
\end{figure}

The entire process of centreline calculation from field quality images is illustrated in Figure~\ref{fig:centrelineCalc}.

\subsection{Prawn Tracking}
\label{subsec:prawntracking}
In an attempt to further enhance the accuracy of calculated prawn lengths, we sought to track individual prawns over multiple frames. Our intention was to estimate the length of each prawn multiple times and then use statistical methods to remove outlier estimates.
The tracking algorithm we used for this purpose is called SORT (A Simple Online and Realtime Tracking algorithm)~\citep{tracking_simple}. SORT uses IoU as its primary metric and applies a Kalman filter~\citep{kalman_filter} to associate a bounding box detected in a captured image (frame) with a bounding box detected on the previously-captured frame. In other words, the algorithm associates images of prawns by examining the overlaps between bounding boxes across frames. 
The performance of the SORT ($max\_age=10, min\_hits=0, iou\_threshold=0.2$) was not satisfactory in our case. This was caused by two major problems:

\begin{itemize}
    \item The first issue is that neither the camera (headset) nor the prawns are stationary. Prawns are fast-moving animals when they jump. In some video sequences, a prawn was lying on the mesh in one frame and jumping (curved up) on the other side of the tray in the next frame. In such scenarios, there is too little overlap between the bounding boxes across frames. As a result, images of the same prawn were treated as images of different animals, i.e., the prawn was not successfully tracked. Adding to this issue, the camera itself is also moving as it is mounted on the head of a human. This can cause tracking to fail even if the prawn is motionless on the tray.
    \item The second problem is the size of the prawn. Small prawns occupy a smaller number of pixels in the frame and subsequently have a smaller bounding box. The IoU metric between small bounding boxes can change dramatically across frames compared to larger bounding boxes.
\end{itemize}

There are several possible ways to obtain a more robust tracking algorithm. For example, we could use the camera's built-in IMUs (Inertial Measurement Unit) to offset camera motion and explore more sophisticated tracking algorithms such as DeepSort~\citep{tracking_deep}. DeepSort is a tracking model that utilises a deep neural network to generate features of the prawns and use the similarities between features to associate prawns across frames. However, one potential challenge is that prawns all look very similar to the human eye. It is unknown if there are sufficient differences/similarities in the features to make such an algorithm work. We are interested in investigating this issue in the future.

\section{Results and Discussion}
\label{sec:results}
\subsection{Prawn detection}
\label{subsec:prawndetectionres}
Table~\ref{tab:detector-benchmarking} shows a brief summary of the top-performing models for prawn segmentation. We used COCO detection evaluation metrics~\citep{coco_eval} to benchmark the models, specifically, the mean Precision (mAP) and mean Average Recall (mAR). In general, precision measures the accuracy of the predictions. i.e. the percentage of correct predictions. Recall refers to the percentage of total relevant results correctly predicted by the model. The precision and recall are calculated using the following equations. TP refers to True Positive ($score \geq 0.50$), FP refers to False Positive, and FN is False Negative.

\begin{equation}
    Precision = \frac{TP}{TP + FP}
\end{equation}
\begin{equation}
    Recall = \frac{TP}{TP + FN}
\end{equation}

In COCO, AP and AR are averaged over multiple Intersection over Union (IoU) from 0.50 to 0.95 with a step size of 0.05, whereas $AP^{IoU=.50}$ is computed at a single IoU of 0.50. The details of COCO evaluation metrics and implementations can be found in~\citet{coco_eval}.

\begin{table}[h]
    \centering
    \caption{A summary of benchmarking results for various detectors.}
    \label{tab:detector-benchmarking}
    \begin{tabular}{lccc}
    \toprule
    Detector & $mAP$ & $mAP^{IoU=.50}$ & $mAR$ \\ 
    \midrule
    ResNet50 + Mask RCNN & .556 & .883 & .619 \\ 
    \midrule
    ResNet101 + Mask RCNN & .552 & .881 & .613 \\ 
    \midrule
    ResNeXt101 + Mask RCNN& .574 & .889 & .639 \\
    ResNeXt101 + Cascade RCNN& .575 & .885 & .639 \\ 
    \midrule
    Swin-Small + Mask RCNN & .545 & .887 & .611 \\
    Swin-Tiny + Mask RCNN & .543 & .892 & .618 \\ 
    \midrule
    Detector ResNet101 + HTC & 0.569 & \textbf{.898} & 0.632 \\ 
    \bottomrule
    \end{tabular}
\end{table}

In our case, we are particularly interested in mAP (IoU=0.5). Thus, HTC with DetectoRS101 as the backbone, which had the best performance on the test dataset, was used in the final prawn length calculation pipeline. A 5-fold validation on the chosen model is shown in Table~\ref{tab:htc-5-fold}.

\begin{table}[h]
    \centering
    \caption{Five-fold validation on HTC + DetectoRS101.}
    \label{tab:htc-5-fold}
    \begin{tabular}{lccc}
    \toprule
    Fold & $mAP$ & $mAP^{IoU=.50}$ & $mAR$ \\ 
    \midrule
    1   & .569 & .898 & .632 \\ 
    2   & .555 & .873 & .620 \\ 
    3   & .581 & .926 & .648 \\
    4   & .573 & .898 & .659 \\ 
    5   & .590 & .920 & .659 \\ 
    \midrule
    Mean    & .574  & .903 & 0.639 \\ 
    \midrule
    SD      & .012  & .019 & 0.013 \\ 
    \bottomrule
    \end{tabular}
\end{table}

\begin{figure}[h]
    \centering
    \includegraphics[width=.9\textwidth]{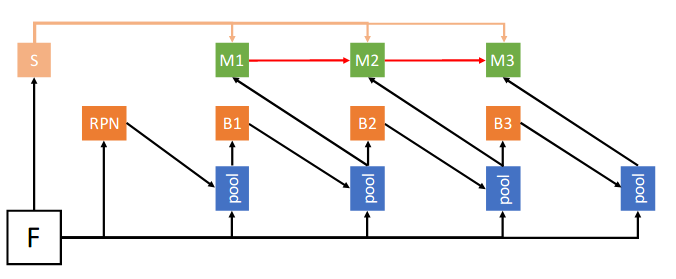}
    \caption{An illustration of the architecture of HTC~\citep{HTC_2019}. M refers to mask (segmentation), and B refers to bounding box (detection).}
    \label{fig:htc}
\end{figure}

Unlike Mask RCNN and Cascade RCNN, HTC (Hybrid Task Cascade) brings improved performance by interweaving the detection and segmentation tasks for joint multi-stage processing and using a fully convolutional branch to provide spatial context (see Figure~\ref{fig:htc}), which helps distinguish foreground from cluttered background~\citep{HTC_2019}. Overall, this framework can learn more discriminative features progressively while integrating complementary features together in each stage.

The backbone, DetectoRS, introduces the Recursive Feature Pyramid (RFP) and Switchable Atrous Convolution (SAC). RFP incorporates extra feedback connections from Feature Pyramid Networks (FPN) into the bottom-up backbone layers, while the SAC, which convolves the features with different atrous rates and gathers the results using switch functions. By combining RFP and SAC, DetectoRS gains significantly improved performance compared to traditional ResNet.

\begin{figure}[h]
	\centering
	\begin{subfigure}{0.4\textwidth}
		\includegraphics[width=\textwidth]{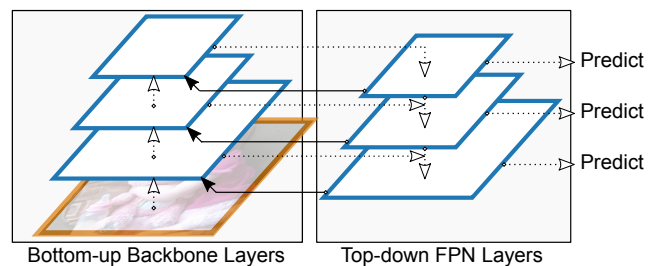}
		\caption{Recursive Feature Pyramid.}
		\label{fig:rfp}
	\end{subfigure}
	\hfill
	\begin{subfigure}{0.55\textwidth}
		\includegraphics[width=\textwidth]{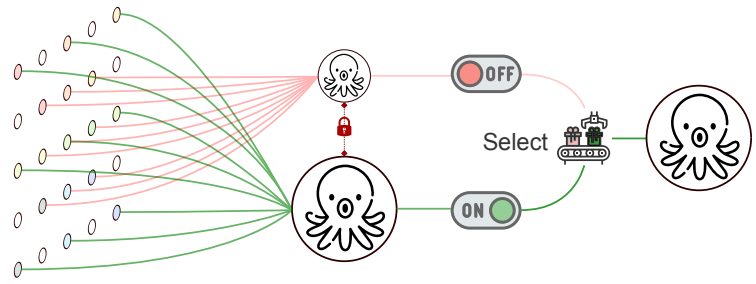}
		\caption{Switchable Atrous Convolution.}
		\label{fig:sac}
	\end{subfigure}
	\caption{An illustration of the Recursive Feature Pyramid and Switchable Atrous Convolution used in DetectoRS~\citep{DetectoRS_2021}.}
\end{figure}

The CV pipeline was then used to process 63 tray check recordings over 23 days. In total, 13,661 prawn instances were detected by the prawn detector across 4067 frames.

\subsection{Prawn growth prediction}
\label{subsec:prawngrowth}
A key research objective was to find out how effective the length measurements are from CV and ML methods when applied to images collected under operational field conditions. Field quality images are normally noisy in nature. Among the images used for validation of CV methods, about two-thirds of the images were discarded because of the poor depth images. 

\begin{figure}[h]
\centering
    \includegraphics[width=.6\textwidth]{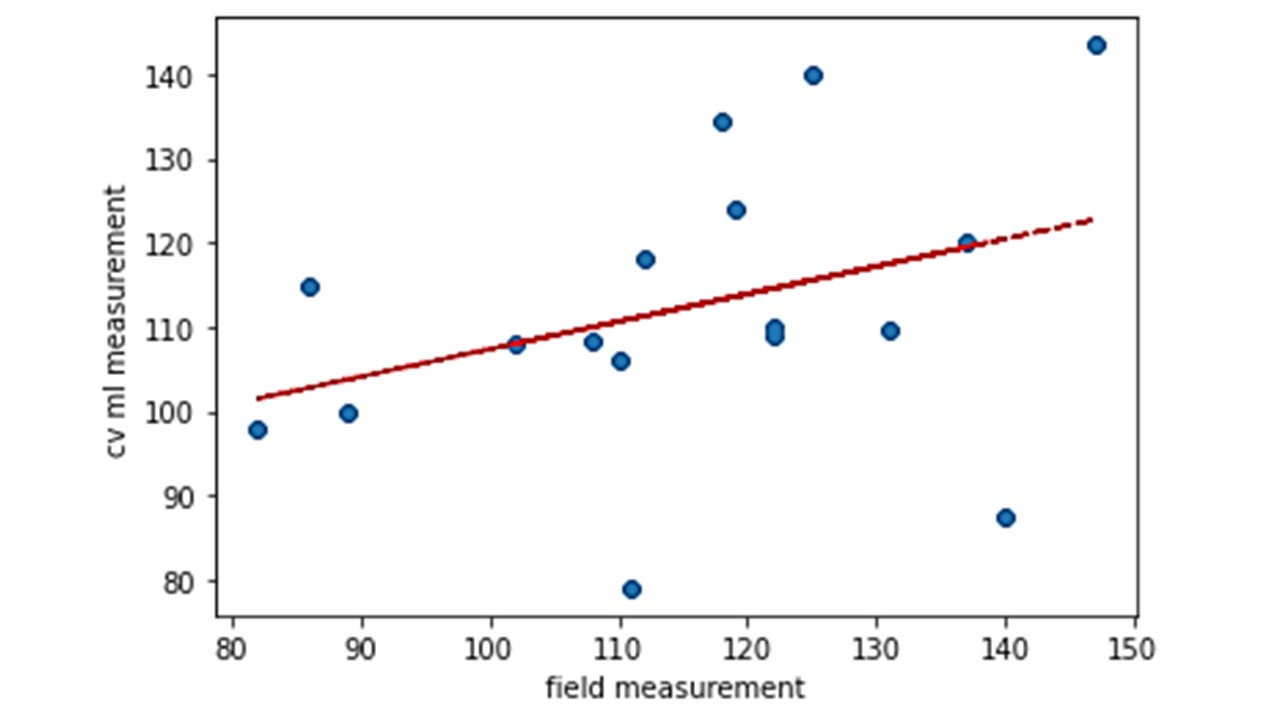}
    \caption{Scatter plot showing the relationship between field measurements and the length measured by CV-ML method. The red line indicates the trendline obtained by fitting the first order polynomial between x and y coordinates.}
    \label{fig:act_vs_pred}
\end{figure}

A scatter plot showing the relationship between field length measurements and the ones computed by the CV-ML method is presented in Figure~\ref{fig:act_vs_pred}. Because of the small sample size (from each feed tray), it's possible to have some outliers as evident from sample measurements around DOC (day of culture) 110 and 140 where the length is very small. Otherwise, the trend line (in red) shows growth over time. Also, the variation in measurement over time is clearly visible in the plot that's unlikely to be visible in the cast netting process. 

We also obtained summary statistics of prawn lengths over time (first four weeks), and the results are presented in Figure~\ref{fig:Trendline}. The top row represents the time series box plot on a different day of culture (DOC). Each box plot represents the summary statistics of prawn length for that day. The first column represents the length estimated using the cast net method. The second column represents the summary statistics of the samples collected from the feed tray during the video recordings (five to six of them), and the third and fourth are the summary statistics obtained using the computer vision method (the third column without tracking and the fourth column with tracking). Following are the key findings from these graphs:

\begin{figure}
\centering
    \includegraphics[width=1.0\textwidth]{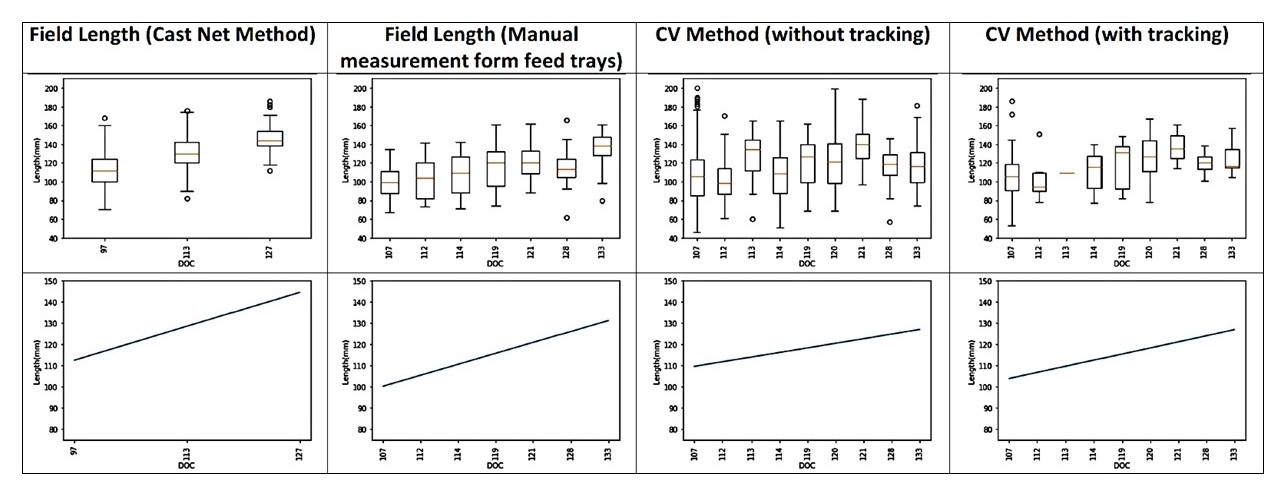}
    \caption{Length of sampled/detected prawns and their variations over time. Top row: time series box plot; Bottom row: linear trend line fitted to median lengths on each day.}
    \label{fig:Trendline}
\end{figure}

\begin{enumerate}
    \item Prawns in samples collected with a cast net (column one) are larger than prawns in samples collected by raising a feed tray (column 2). We need to find a way to bridge this gap.
    \item The trendline for each length measurement method shows an upward trend indicating that prawns are growing over time. While the rate of growth is not exactly the same, the CV method trendlines (column three) are similar to trendlines based on feed tray samples (column two)
    \item The trendline based on tracking-based CV methods (column 4) is closer to that measured directly from feed tray samples than the trendline from CV methods without tracking. This demonstrates that tracking was effective to some extent
    \item The boxplots for each day show the variation of length within single samples of prawns. For the day and captured by all methods. The box plots in the cast net method show an upwards trend of median only. However, high sampling methods (columns two, three, and four) show significant size variation over time. This is aligned with the observation from the project.
\end{enumerate}

\subsection{Prawn distribution}
\begin{figure}
\centering
    \includegraphics[width=1.0\textwidth]{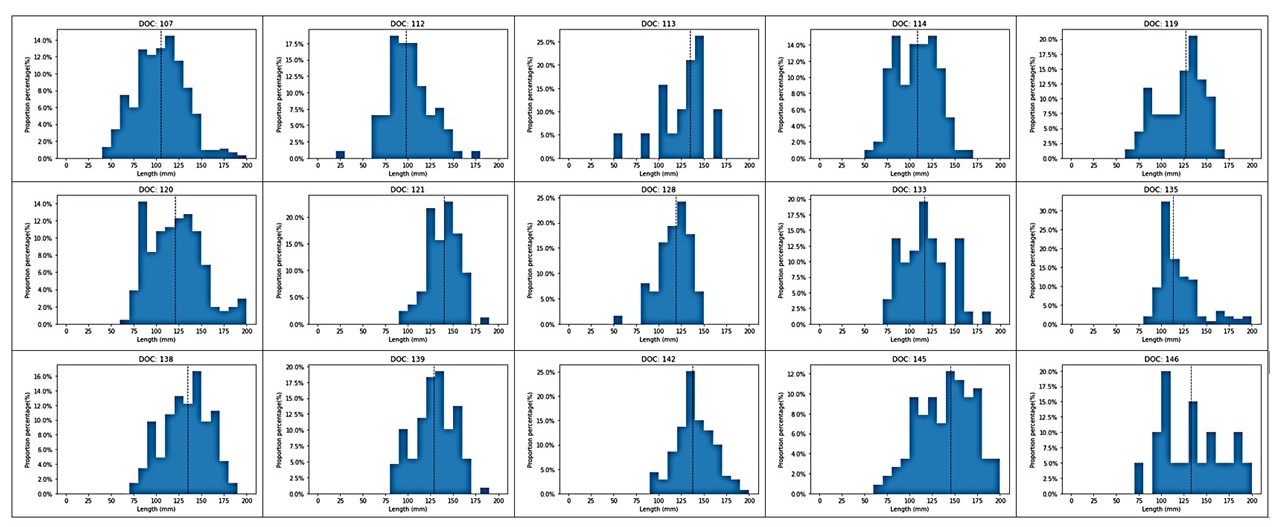}
    \caption{Length distribution across DOC. Note how the median length (black dotted line) is shifting towards the right as DOC becomes higher}
    \label{fig:Length-Distribution}
\end{figure}

This section describes the distribution of prawns in the dataset over time. Figure~\ref{fig:Length-Distribution} shows the distribution of prawns at different DOCs (Day of Culture). The median line in each subplot is presented using a black dotted line. As DOC increases, the median line moves to the right of the plots implying prawn size increases over time. Note that the distributions between successive DOCs may look a bit inconsistent because of samples that were pulled by the feed tray. It's not unlikely given the small sample size. However, if we look at the global trend (median line), the growth is visible. Also, the variability between samples is very evident from the plots, and it's only possible because of high-frequency sampling.

\section{Conclusion}
\label{sec:conclusion}
In this paper, we present an approach that measures prawn size automatically during feed tray inspection using a smart headset, google glass, computer vision and machine learning method. The smart headset with the combination of google glass does not add any extra work for farmers but is capable of capturing images (both colour and depth). The deep learning-based computer vision method then detects the prawns, and the corresponding depth camera segment is used to estimate the length of the prawns. The distribution of prawn length and growth trend over the growing season, as computed by our approach matches closely with the field measurements. In future, we aim to utilise this approach for phenotype measurement in livestock and crops.

\section*{Acknowledgements}
The authors would like to thank CSIRO’s Digiscape Future Science Platform for funding this project.

\bibliographystyle{elsarticle-harv} 
\bibliography{AE_2022_Xi}

\end{document}